\documentclass[preprint,12pt]{elsarticle}

\usepackage{amssymb}
\journal{Cognitive Systems Research}

\begin{document}

\begin{frontmatter}



\title{What you need to know about a learning robot: Identifying the enabling architecture of complex systems}

\author[label1]{Helen Beierling \corref{cor1}}
\author[label1]{Phillip Richter} 
\author[label1]{Mara Brandt}
\author[label2]{Lutz Terfloth}
\author[label2]{Carsten Schulte}
\author[label3]{Heiko Wersing}
\author[label1]{Anna-Lisa Vollmer}
\cortext[cor1]{helen.beierling@uni-bielefeld.de}
\affiliation[label1]{organization={Bielefeld University, Medical School EWL, Interactive Robotics},
            addressline={Universitätsstraße 25},
            city={33615 Bielefeld},
            %postcode={33615},
            %state={Nordrhein-Westfalen},
            country={Germany}}

\affiliation[label2]{organization={Paderborn University, Faculty of Computer Science, Electrical Engineering and Mathematics, Didactics of Informatics},
            addressline={Warburger Straße 100},
            city={33098 Paderborn},
            %postcode={33098},
            %state={Nordrhein-Westfalen},
            country={Germany}}

\affiliation[label3]{organization={Honda Research Institute Europe GmbH},
            addressline={Carl-Legien-Str. 30},
            city={63073 Offenbach am Main},
            %postcode={},
            country={Germany}}

% \author{}

% \affiliation{organization={},%Department and Organization

\begin{abstract}
Nowadays, we are dealing more and more with robots and AI in everyday life. 
However, their behavior is not always apparent to most lay users, especially in error situations. 
As a result, there can be misconceptions about the behavior of the technologies in use. 
This, in turn, can lead to misuse and rejection by users. 
Explanation, for example, through transparency, can address these misconceptions. 
However, it would be confusing and overwhelming for users if the entire software or hardware was explained.
Therefore, this paper looks at the `enabling' architecture.
It describes those aspects of a robotic system that might need to be explained to enable someone to use the technology effectively.
Furthermore, this paper is concerned with the `explanandum', which is the corresponding misunderstanding or missing concepts of the enabling architecture that needs to be clarified.\\
We have thus developed and present an approach for determining this `enabling' architecture and the resulting `explanandum' of complex technologies.
\end{abstract}

\begin{graphicalabstract}
\vspace{0.5pt}
\centering
\includegraphics[width = \textwidth]{figures/FlussDiagrammKomplett}
\end{graphicalabstract}

\begin{highlights}
\item Research highlight 1 : Researching the Applicability of Didactic Reconstruction to XAI in HRI.

\item Research highlight 2: Four-step model to identify an enabling architecture and a corresponding explanandum.

\item Research highlight 3: Application of this approach to identify an enabling architecture to an example.

\end{highlights}

\begin{keyword}
robotics \sep HRI \sep explainability \sep didactics \sep didactic reconstruction


\end{keyword}

\end{frontmatter}


\section{Introduction}
% Alltags verwendeten Roboter 
In our everyday lives, we get in touch with more and more robots. 
For example, Paro by \citet{paro} in the care sector, DaVinci by \citet{davinci} in the operating room, temi by \citet{temi} in hospitals, Pepper by \citet{pepper} in retail, and iRobot vacuum cleaners by \citet{irobot} in our homes.
%Roboter raussuchen
To only name a few.
%They are used in environments where the task the robots need to perform can hardly be learned in advance since they are heavily based on the user's preference.
If we have robots in our immediate environment that support us in everyday tasks, then these should be adapted to our personal preferences.
Therefore, the users must train the robots to customize them to their needs.
This is done by training during a human-robot interaction. 
One common and effective method to perform such training is by interactive task learning \citep{thomaz2006reinforcement}.
This method uses the user's feedback for the learning process.
For example, directly by using human feedback as a reward 
%Zitat
or by demonstration. Different approaches for this were, for example, presented by \citet{thomaz2008teachable}, \citet{akalin_reinforcement_2021}, and \citet{argall2009survey}.
Those methods are thus dependent on the informativeness of the user's feedback.
They can suffer if the users have misconceptions within their mental model of the inner functioning of the robot. This phenomenon was examined by \citet{hindemith2020robots}.
For example, if the robot has human features like eyes and is taught by demonstration when the robot does not know what the user wants or expects, the user tends to do extra movements so that the robot can follow better. \citet{vollmer2014robots} found this movement behavior in their studies. These movements might then be learned as part of the target movement, slowing down the learning process.
% This can be problematic since the robot may interpret the speed as relevant and learn it, which may worsen the situation.
Another example of these misconceptions in learning based on direct reward is the misuse of the reward for motivation, which overweights the behavior graded, which was shown by \citet{thomaz2008teachable}.
A natural approach to resolve those misconceptions is by explaining the inner lying architecture of the robot \citet{thomaz2008teachable}.
However, these inner structures of the robot can be very complex.
Furthermore, the users are not willing to invest a lot of time into the training of the robot. 
Therefore, the best option would be only to explain those parts of the robot's architecture or algorithm that the user misunderstood or is missing in its mental model such that the training is not frustrating and is successful as quickly as possible.
This was indicated in the paper by \citet{LukasAnna}.\\
Hence, a structured procedure is needed to identify these parts of the architecture.
In this paper, we will present our approach and its development.
First, we will introduce the related work in the field of explainability in human-robot interaction and mental models.
We have worked out our approach using a robot and algorithm architecture as an example.
We will present this example architecture in section \ref{sec:ExampleArchitecture}. 
Afterward, we will present how we worked out this approach and the application of the elaborated steps to our exemplary architecture in section \ref{sec:methods}.
Lastly, we will summarize and discuss our approach and present future work.

\section{Related Work and Example Architecture}
In this section, we will first provide an overview of the field of research. 
% We are focusing on mental models in general and in robotics as well as their evaluation.
% Lastly, the field of interactive task learning in robotics is covered.
Afterward, we will present our example architecture used throughout the paper to illustrate the steps of the presented approach.
% Hier sagen das wir uns stark auf missconception in mentalen model fokussieren und daher gehen wir erst auf unser verständnis des Begriffes ein und ordnen ihn im Feld Robotic und HRI ein 
% Dann gehen wir auf weitere XAI methoden ein`
% am ende noch eine übersicht über mögliche andere Architekturen um klar zu machen das unsere ok ist 
%so that we first look at other architectures. 
%Evt andere XAI methoden
% Then we will look at related work in the area of mental models, especially with a focus on human-robot interaction.  
% Lastly, we will present different approaches for explainability in interactive robot learning with humans in the learning loop.
%Structure anpassen
\subsection{Related Work}
%Intro
This paper aims to explore the field of misconceptions in human mental models and their impact on explainable artificial intelligence. 
For this reason, the concept of a human's mental model is first introduced, and the context of human-robot interaction is discussed. 
Thereupon, the representation and acquisition of a human mental model is further elaborated. 
An introduction to the concept of interactive task learning in robotics follows this.
Finally, several approaches to explainability in interactive robot learning by humans are presented.\\
%Mental Models
Humans build an internal mental model of their environment to understand and interact with their surroundings. These cognitive representations are individual mental constructs that evolved from past impressions and form a human's view of himself and the world around him as described by \citet{gentner2014mental}. \\
%Mental Models of Robots
Therefore, humans also build internal mental models of technical systems they interact with. A human's mental model of a robot's capabilities is influenced by the robot's social cues, humanlike movement, and anthropomorphism, which was shown by \citet{kiesler2005fostering} and due to many years of misleading portrayals of robots in the media, humans might develop false assumptions about the capabilities of robots as indicated by \citet{bruckenberger2013good}.\\
%Mental Models towards HRI
This also impacts human-robot interaction since previous research has shown how humans form different mental models towards different types of robots \citep{lee2005human}. Correspondingly, human behavior is also different towards the respective robot \citep{mumm2011human}.
Further, the increasingly human-like design of robots may lead to the assumption of human-like capabilities and learning methods. \\
All of this contributes to the development of false human expectations about robots' real functioning and behavior. This trend was also discussed by \citet{berzuk2023clarifying}.
This expectation discrepancy finally results in human disappointment, frustration, and failure in human-robot interaction. This effect was, for example, described by  \citet{berzuk2023clarifying} and \citet{malle2020trust}.\\
%Recieving a MM of Robot
The task of representing a human's mental model of a robot is not trivial and has been explored in various research for example by \citet{kiesler2002mental}, \citet{lee2005human}, and \citet{powers2006advisor}. 
Such a mental model could be queried by asking a human about a robot. However, this method by \citet{brooks2019building} is very intrusive and involves the risk of failure due to a lack of expertise. 
However, previous research by \citet{perez2015fast} and \citet{brooks2019building} has revealed that a human's low-level actions toward a robot reflect information about the human's mental model of the robot and, therefore, could be used to create a representation of a human mental model. \\
%% umformulieren
%ITL HRI
Interactive task learning is a teaching method in which an agent learns from a human by receiving immediate feedback from the human while performing a task. The agent then can adjust its behavior accordingly and thus improve its performance in this environment as described by \citet{kirk2014interactive} and \citet{ramaraj2021unpacking}. There are several interactive teaching methods: A commonly practiced approach is teaching by demonstration, in which a task is executed by a human and then imitated by an agent. For example, this type of approach was listed by \citet{argall2009survey}. Another interactive teaching method is learning by advice. Such an approach was, for example, used by \citet{kuhlmann2004guiding}. Alternatively learning by rewarding feedback can be used as described by \citet{knox2009interactively} and \citet{thomaz2006reinforcement}.
However, for all interactive learning algorithms, research has shown how humans adapt their teaching approach as they develop a mental model of a robot's capabilities, limitations, and how it learns. This was shown in various studies, among others, by \citet{thomaz2008teachable}, \citet{LukasAnna}, and \citet{vollmer2009people}.

%selbes paper hat gezeigt das das feedback einen positiven bias hat, vielleicht wichtig je nach study ergebnis

%HRI Explainability
% papaer was explizit sagt war nicht nur asudrucklich input ouptu verhahlten,
%unser problem nicht nur input output verhalten
%anderes problem ganzen algorithmus zu erklären ist nzu komplex und dauert zu lang

% Soll hier die recherche hin? 
% Was gibt es an relvanter Architektur
\subsection{Example Architecture}
\label{sec:ExampleArchitecture}
In our setting, we used probabilistic movement primitives (ProMP) developed by \citet{paraschos2013probabilistic} to represent the robot's movements. 
They are very efficient in representing complex movements and are a commonly used tool in robotics.
This efficiency is based on the fact that ProMPs decode the movement as parameters of the distribution of movements, and the number of PI\textsuperscript{BB} parameters is smaller than the information needed to save the whole trajectory.
ProMPs are also more flexible than directly saving the trajectory. 
We used the learning algorithm PI\textsuperscript{BB} presented by \citet{stulp:hal-00738463}, which is a black-box optimization algorithm. 
The approach works with the probabilistic nature of ProMP and can optimize those based on user feedback. 
We used the robot arm Kinova Jaco 2 \citet{kinova}.
It has six degrees of freedom and is controlled based on robot operating system (ROS) version noetic by \citet{noetic} and python 3 by the \citet{python}.
Since ProMPs serve in our context more as a datatype than as part of the architecture suffering from lousy feedback, we have limited the example architecture to PI\textsuperscript{BB} and the robot.
\section{Methods}
\label{sec:methods}
In this section, we will describe the development and application of our approach.
\paragraph{Didactic Reconstruction and Reduction}
Didactics is an intuitive field of research to turn to.
Among other things, Didactic provides structured methods to reduce complex topics, structure them and present the result to lay people of the topic.
% Those include a method for identifying the enabling architecture and the explanandum. 
% Here besser beschreiben
We exchanged with experts from the field and reviewed according literature.
Based on this, we selected didactic reconstruction and reduction as the foundation of our approach.
These methods allow us to identify the enabling architecture and the explanandum.\\
The didactic reconstruction was designed by \citet{kattmann1997modell} in 1997 and has been developed further since then.
It addresses the problem of conveying complex information with the layperson's knowledge in mind. 
Instead of reducing complex information to the lay user, it starts with the lay user and builds on their prior knowledge.
This is important since the human mental model is also formed on a base domain of prior knowledge, as shown by \citet{staggers1993mental}. 
A human's perception of a task he has to perform or learn is highly influenced by the concepts that have been built from prior knowledge. 
This observation was among others found by \citet{norman2014some}; 
it can be assumed that the mental model of a human, in the context of didactic reconstruction, corresponds to the prior knowledge of a lay user before a teaching scenario.\\ 
The didactic reconstruction considers three aspects: the information to teach, the pupil's mental models, and the form of presentation used to convey the information. 
First, the approach takes an overview of the factual situation of the selected information to teach. 
Afterward, the prior experience and knowledge of the target audience are determined.
Last but not least, the form of presentation that conveys the information best and fits the target audience is specified. 
This allows to extend the pupil's conceptions and adapt and classify them in the factual situation.
The identified content significantly influences the determination of the presentation form.
Thus, the enabling architecture needs to be identified first. 
Therefore, the form of presentation is future work for this paper. 
Hence, we focus on three aspects: the factual situation, the pupil's mental models, and the synthesis of the two.
For the factual situation, we queried experts of our example architecture. 
As the literature suggests, the raw expert explanation is too complex for laypersons, even though they had them as the target audience. 
This discrepancy of given explanintions and needed explanations was found in different papers for example by \citet{lehner2020didaktische},  \citet{mietzel2007padagogische}, and \citet{bromme1998verstandigung}.
We, therefore, need to reduce those explanations.
The reduction process was based on the second concept, didactic reduction. 
The approach is similar but older than the reconstruction and focuses solely on reducing information.
It provides more refined reduction techniques. 
Therefore, we used it as a base to reduce the expert explanation to find the aspects that the experts included and thus consider necessary for lay users to understand the architecture. 
The reduction also eases the synthesis of the factual overview of the expert's and the pupil's mental models since the overlap of aspects, and especially the missing concepts, can be systematically evaluated.
% Schlägt qualitative Inhaltsanlyse vor ist auch ein anwendbares mittel der didaktischen reduktion dar das anwendbar ist
% Schritte Mappen zu unsrem vorgehen um in den Kapiteln aufzugreifen und am ende zusammenzufassen
\\
The didactic reduction was developed from 1958 to 1991 by an interaction of various authors who complimented and continued each other's work.
Central names here are Dietrich Hering and Gustav Grüner.
We used the more recent work by \citet{lehner2020didaktische} as the base for the reduction of this paper.
%Zitat und ist das wirklich gut Lutz fragen
Didactic reduction occurs when extensive and complex issues are processed to make them clear and comprehensible for the learners,
% It is described as a process that makes broad and complex topics clear and comprehensible,
as described on p.9 of \citet{lehner2020didaktische}.
In contrast to the didactic reconstruction, it is based on the factual situation and reduces it without considering the pupils from the beginning.
% Vorschlag: Didactic reductio is described as a process that makes broad and complex topics clear and comprehensible for the learners, p.9 \citet{lehner2020didaktische}.
Regarding the application of didactic reduction, there are different types and focuses.
Firstly, there is a distinction between horizontal and vertical reduction.
Horizontal reduction aims to reduce the form of presentation. 
This can be done by visualizations, for example. 
Vertical reductions intend to reduce the content that is presented.
The explanation scope of the expert explanations was not very high on average, but their complexity was. 
Thus, we want to perform a vertical reduction of the explanation.
Another distinction is made in the didactic reduction between quantitative and qualitative. 
Quantitative reduction reduces the amount of information.
Qualitative reduction reduces the complexity of the content.
% Also, here the focus of the reduction is on reducing the complexity, then reducing the amount of content.
Thus, we focus on vertical qualitative reduction. 
However, most of the suggested reduction techniques for vertical qualitative reduction do not apply to our scenario since they are based on everyday examples of the phenomenon to be explained. 
But the interaction with robots does not occur every day for most users. 
We also focus on systematic methods that are as little as possible biased and are reproducible since the approach should also apply to other architectures. 
One option that fulfills these criteria is the systematic creation of a concept map. 
This is one approach suggested by \citet{lehner2020didaktische} on p.155 as a didactic reduction method, and the didactic reconstruction also supports it. 
% The didactic reconstruction recommends a qualitative content analysis to clarify the factual situation, from which a concept map can visualize the results. 
%Thus, we conducted a qualitative content analysis of the expert explanation and visualized the results within a concept map.  
To create the concept map systematically, we defined methodical analysis steps to determine the concepts and interactions for the final concept map.
\\
First, we queried architecture experts to get our example architecture that forms the concept map's base. 
Since we are interested in architecture in the context of explanation, we needed an explanation of the algorithms as a basis for the analysis.
On the one hand, those explanations would already be reduced since the explainer would need to start at some point. 
On the other hand, the explanation will be structured such that the analysis will provide us with an order. 
To sum up, we conducted a study where we queried experts to explain PI\textsuperscript{BB} with lay users as the stated target audience.
\subsection{Factual Situation - Expert Study}
\label{sec:FactualSituation}
Before we go into detail regarding the study, we want to make clear what we define as an expert.
\paragraph{The Experts}
As an expert, we define the scientists that worked on the initial paper of PI\textsuperscript{BB} or were referenced as related work in those papers. 
We contacted over 60 experts via mail and got 19 responses.
Of those 19 experts, only 9 gave evaluable and complete feedback.
All nine experts worked recently or during a long period of their careers with the algorithms.
\paragraph{The Study}
Since most experts are active in industry or research, they are busy, so appointment finding is difficult.
Additionally, the experts are located worldwide, which adds to the difficulty of different time zones. 
We thus chose the questionnaire over a direct interview. 
This also allows the experts to fill out the questionnaire in multiple sessions.
The questionnaire was divided into two parts.  \\
In the first part of the questionnaire, the expert stated their state of knowledge with our example architecture.
This included the amount of experience they have had with the algorithm in their careers and when they last worked with it. 
All experts have worked a lot with the algorithm until today or, to a lesser extent, in the recent past.
They were also asked to grade their knowledge of the algorithm and its complexity from their perspective.
On average, the experts evaluated their knowledge and the algorithm's complexity as medium.
%In the second section of the questionnaire, we asked about keywords and an explanation for our example algorithm. 
In the questionnaire's second section, we asked for an explanation of our example algorithm. 
% The keywords were queried first.
% We were interested in those since they show where the authors categorize the algorithm in the field of machine learning and robotics, and it reveals their perceived main features of the algorithm. 
%Afterward,
We queried explanations targeted to lay users.
Besides the context, we stated that the scope of explanation could be individually defined.
This means the experts can give bullet points or full-text answers.
As already mentioned, the raw expert explanation was not helpful for laypersons.
We presented the explanations to lay users.
Nearly all the explanations contained unknown concepts and interactions that were not explained nor understood by those lay users.
This might also be connected to the fact that the experts assess the algorithm's complexity primarily as medium from their perspective.
Thus, the raw expert explanations are not the enabling architecture. 
Even if they are already reduced, in the sense that the experts explained not all possible information and provided a structure for presenting concepts and interactions. 
Those explanations are the architecture that must be reduced in the following steps. 
Regarding the didactic reconstruction, the expert statements form the basis for the factual situation to determine the architecture of our example algorithm. 
\subsection{Reduction}
\label{sec:architectureReduction}
\begin{figure}
    \centering
    \includegraphics[width= \textwidth]{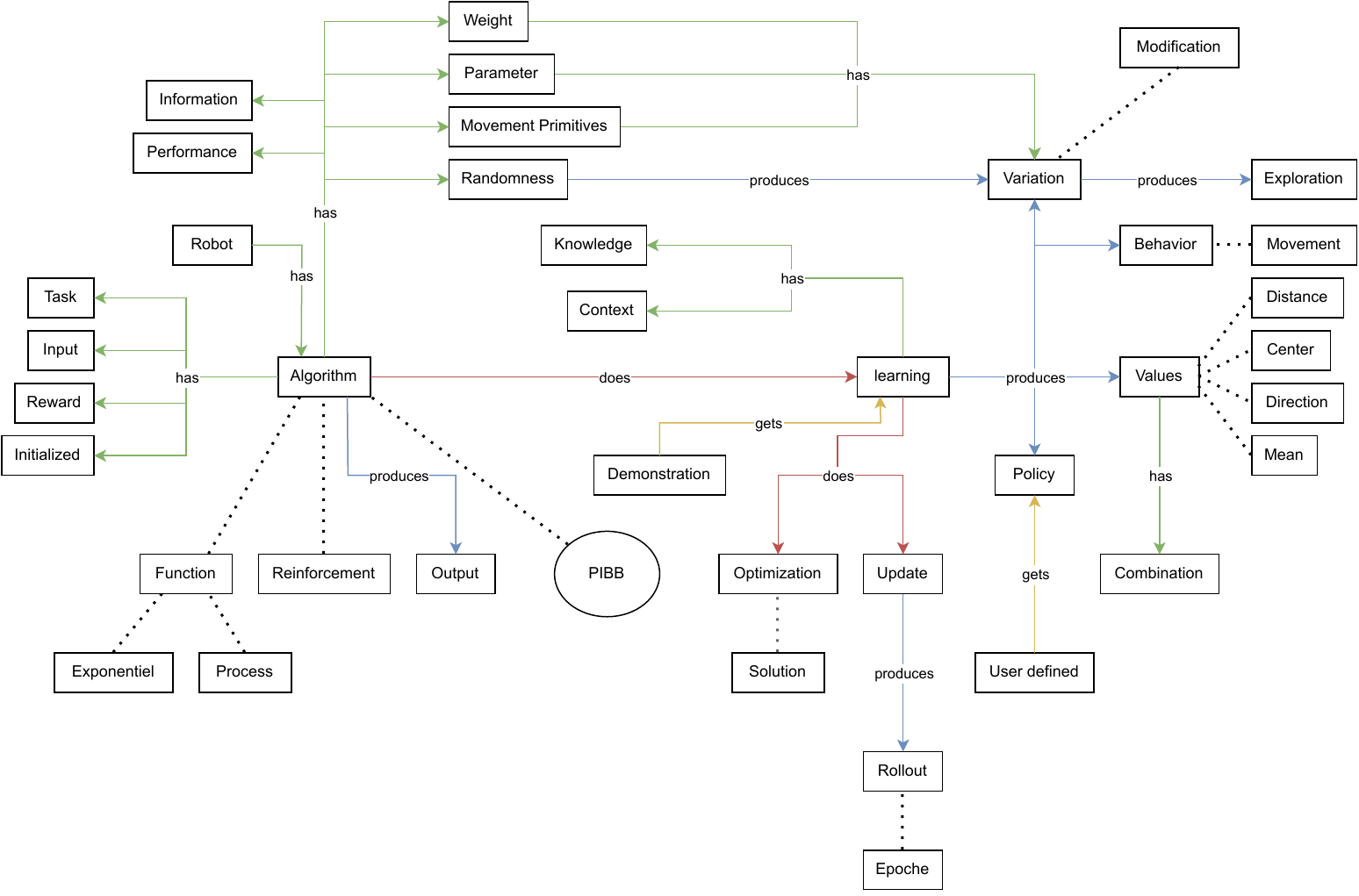}
    \caption{The concept map of PI\textsuperscript{BB}, based on the analysis and reduction of the expert explanation.}
    \label{fig:ExpertRelations}
\end{figure}

With the expert responses, we have an architecture to work with. 
% The goal is to identify what, needs to be explained of an algorithm to enable lay user to interact more effective with robots.
The concept map creation can then be based on the results of the analysis steps described in the following. 
\\
First, we preprocessed the experts' explanations by removing all function words.
Afterward, we analyzed the emerging concepts and interactions quantitatively by determining their frequency.
This included the number of times each expert mentioned a concept.
It was also noted how frequently all experts mentioned the concepts and interactions.
The difference is that the first frequency does not represent whether one author mentions a concept very often. 
The second considers the overall frequency of the concepts regarding all experts.
For instance, if a concept or interaction was mentioned frequently by one expert and never by the others, this was documented.
After the quantitative analysis, we merged concepts and interactions that are identical in meaning.
This, for example, includes merging plural and singular words as well as different grammatical cases. 
After that, we started with the post-processing of this analysis results. 
This includes merging interactions and concepts that are generally similar in meaning.
An example would be motion and movement, which both describe the same concept.
% Nochmal Angela fragen wie das heißt 
A less forthright way of merging is only to merge concepts and interactions, which mean the same in the context of the explanation.
For example, behavior and action refer to the same concept for PI\textsuperscript{BB} because they both describe the base for optimization of the algorithm.
In those cases, we kept the concept or interaction close to our setting.
We kept the more abstract formulation for concepts and interactions not represented in our setting. 
When merging, we also updated the frequency of the concepts.  
The last post-processing step is reducing the concepts and interactions based on frequency.
Concepts and interactions mentioned less than three times in total were dropped. 
Furthermore, we only kept concepts and interactions mentioned by at least two experts.\\
For the visualization, we used a concepts map as the didactic reduction suggested as one possible approach for the reduction. 
The creation of the concept map, based on the systemic analysis, was also done systematically. 
The nodes of the concept map represent the reduced concepts of the experts.
For the connections, we defined relations: “has”, “gets”, “produces”, and “does”.
Furthermore, we added a dashed line to indicate that the concept is hierarchically part of another concept. 
For example, the apple is part of the concept of fruit. 
The preprocessing of the expert interviews, analysis and post-processing, and the described relations resulted in the following concept map (Figure \ref{fig:ExpertRelations}) based on the explanation of PI\textsuperscript{BB}. 
% The first is based on the explanation of PI\textsuperscript{BB}.
This concludes the analysis of the expert interviews.
With this reduction, we can proceed with the didactic reconstruction.
The next step for this is to identify the pupil's mental model. 
In our case, these are the mental models of the users. 
Therefore, we conducted another study to identify their mental models of our learning algorithm PI\textsuperscript{BB}.
This allows us to synthesize the reduced expert architecture with the user's mental models, as this is the next step in the didactic reconstruction. 

\subsection{Pupil's Mental Model - Lay User Study}
\label{sec:LayMentalModel}
\begin{figure}
    \centering
    \includegraphics[width = 0.5 \textwidth]{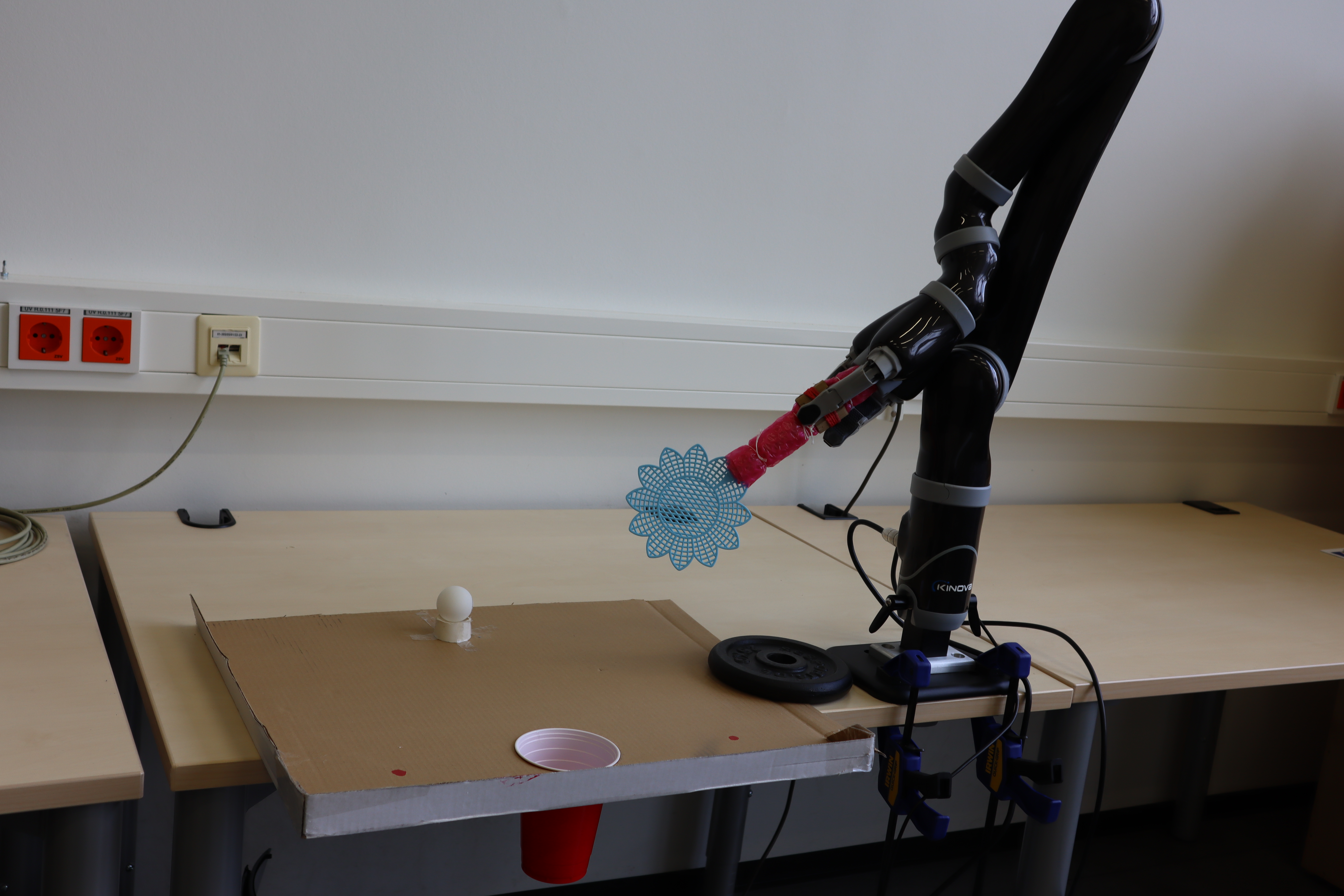}
    \caption{Our example setup. The Kinova Jaco 2 with an obstacle-free self-made golf course. The ball is a light ping-pong ball.}
    \label{fig:set-up}
\end{figure}

\begin{figure}
    \centering
    \includegraphics[width=0.5 \textwidth]{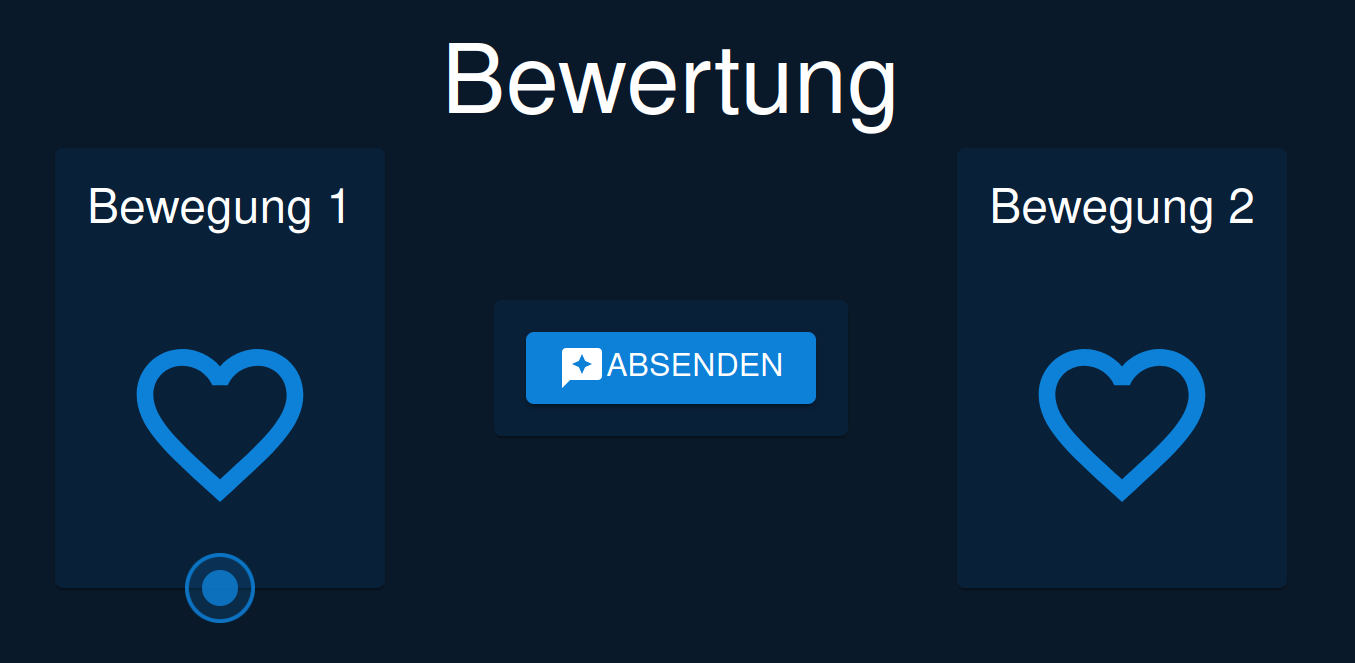}
    \caption{Our example graphical user interface. The title says "\textit{Rating}", and both hearts are titled "\textit{Movement 1}" and \textit{"Movement 2}". The central button is a submit button for the rating and is labeled "\textit{Submit}".}
    \label{fig:GUI}
\end{figure}

We need to assess the users' mental models in the three phases before, during, and after interacting with the robot.
During the interaction, the initial mental model changes continuously \citet{thomaz2008teachable}. 
Thus, the prior and final mental models and transformations are interesting. 
Our approach to this is to ask the user for their mental model of a robot learning situation.
Then let them interact with our example setup and perform a video recall of that interaction afterward.
During the recall, we probe the users if they directly or indirectly mention an expert concept.
This probe allows us to clarify that the user refers to the same concepts as the experts besides naming it the same way.
Furthermore, it eases the synthesis of expert concepts and users' mental models.
Another advantage of the recall technique is that it allows the user to focus more on the architecture since they are less involved after the interaction. 
On the one hand, the mental load of the users is lower, as they do not have to act and describe their actions. 
On the other hand, users not only describe what they do but can also describe the causality of their actions or even comment on the underlying system.
Since the experts explained the algorithm without a specific task or modality in mind, this reduced focus on these two aspects during the video recall also eases the synthesis.
% Thus, we decided that the user should not interact with the robot themselves and instead watch a video of an interaction.
% This is because they tend to describe their actions or focus on the reasons they had for the rewards.
% Both would not focus on the causality between reward and resulting action.
% During the video, the rewards are given by an external source such that the users focus less on the rewarding process and more on the learning itself. 
Afterward, we again query their mental model of a learning situation of a robot.
\paragraph{The Lay User}
This paper defines a lay user as a person with no technical background in robotics and artificial intelligence. 
%For our study, the lay user we recruited were .
In total, we had ten participants to match a similar extent as the experts. 
%Hier nochmal das Paper raussuchen wo der Unterschied zwischen den Gender bei der Interaktion nachgucken 
% Wir werten das nicht aus wollen wir das dann?
% We questioned this because it has already been established that there are differences between the sexes when interacting with robots. On average, women were nicer to robots than men. This could be relevant as it could also influence rewarding behavior. 
\paragraph{The Study}
% Participants were brought to the labratory in such a way that there was as little contact as possible with the robots in the other labs and exhibits.
At first, we informed the participants that the study was concerned with robot training.
Furthermore, we stated that we want to evaluate how users with little knowledge about the system and algorithm use them. 
We also informed them about the remuneration for the study.
We communicated to the users that the data is processed and collected their consent. 
As reasoned above, we then asked them how they imagine how a robot is trained for a task.
We avoided the word `learning` not to bias the participants. 
% Hier einmal schreiben was dabei ca rauskam
The participants indicated various approaches in the initial question on how to train a robot for a task.
After completing the training task, the participants were asked again how they imagined the robot would be trained for a task. 
There were still different approaches mentioned, but the answers were much more similar to the concept of our training scenario.
Furthermore, we asked them about their background in robotics. 
%Half of the participants rated their knowledge as very low and the most of the rest rated it as medium.
On average, the knowledge was graded as low as desired for the study. 
The knowledge was never graded higher than medium on a five-point scale from very low to very high. 
Also, we asked about their educational level.
The participants had a diverse level of education.  
It ranged from having finished school up to having graduated from university. Half of the participants have graduated from university. 
This helps to assess the prior mental model of the participants.
Lastly, we asked for gender because other studies, for example, regarding anthropomorphism, showed differences between the genders regarding their interaction with robots \citet{de2015robots} \citet{green2008sensitivity}. 
Regarding gender, 3 of the participants were female and 7 male. 
After answering the above questions, the questionnaire was locked with a PIN so that participants would not see the final questions before the end of the study. 
Then we showed them our example setup as shown in Figure \ref{fig:set-up} and explained to them that this is the robot they will be working with.
As the task, we used a miniature golf scenario With a track without obstacles (Figure \ref{fig:set-up}).
%Vorschlag: The task is to hit the ball over a short course which does not contain any obstacles, as seen in Figure \ref{fig:set-up}.
The task is easy for users to evaluate and is not too abstract that it distracts from learning. 
It is also interesting enough that we can keep the users' attention.
We stated that they must train the robot to hit the hole in one try.
Furthermore, we informed them that the camera installed will record the interaction but is not used by the robot for the training.
%Vorschlag: Furthermore, we informed them that the camera installed will record for later interview, but is not used by the robot in the study environment.
Then we explained to them that the training is based on their ratings.
After that, we presented our example graphical user interface and explained how they could use it to grade the robot c.f. Figure \ref{fig:GUI}.
We stated that the robot will show them two movements, and they could rate them by liking a movement.
This allows the user to grade both movements as good or both as bad or one as good and the other as bad. 
% The reward was given based on stating the preference between two presented actions.
% Actions that were preferred are, and the action that was not favored is . 
% Users can also state that they preferred none of the actions or both and are graded as.
% Furthermore, we informed them that the task is done when the robot succeeds the first time.
We performed a semi-structured video recall with the participants when the users were finished.
We brought them to a separate table to show them their interaction.
Then they were shown loops of learning.
One loop consists of the movement that was presented and the reward the participants gave for that movement.
Those loops were shown one after another, and the users stated their thoughts regarding the loops.
The participants were informed that the interview will be recorded.
We instructed them that they were supposed to explain to us why they were acting as they did in the video.
Also, we informed them that we would not judge their reasoning and that it is not essential that their statements are correct, and that it is most important that they answer the questions as honestly as possible. 
When they mention a concept stated by the experts or a related concept, we probe them into what it means, as described above. 
Care was taken to ensure that the participants were not given a predefined answer.
% This interview was recorded as well.
After the interview, We asked them how they imagined robots learning as in the beginning, to evaluate the change in their mental model.
% Beschreiben und analysiren was die Nutzer gesagt haben 
Finally, we thanked the participants for their participation and sent them off.
We analyzed the results by using the same analysis as for the expert responses.
The concepts and interactions were counted and reduced as described in section \ref{sec:architectureReduction}. 
\begin{figure}
    \centering
    \includegraphics[width = 1\textwidth]{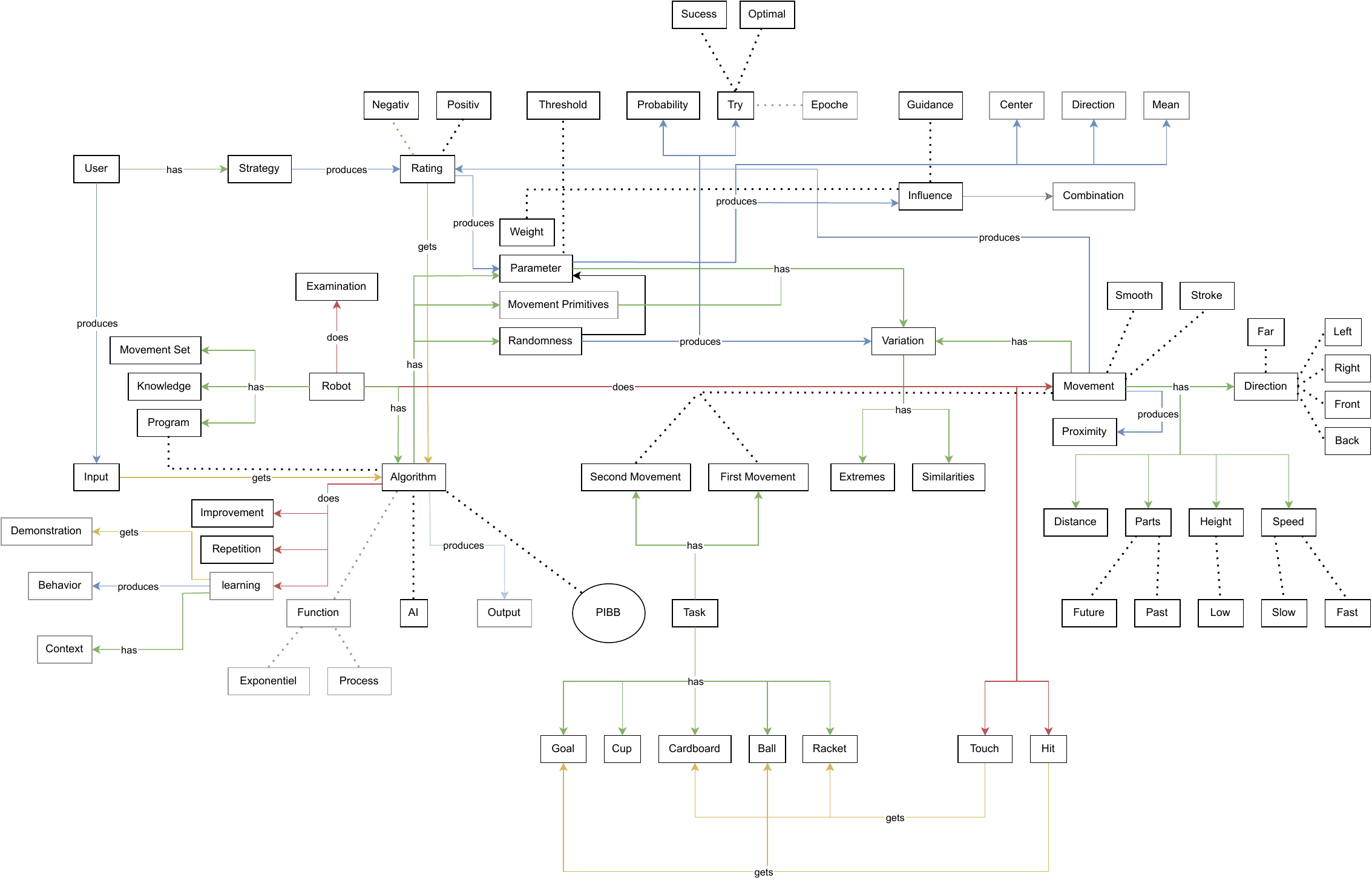}
    \caption{The concept map of PI\textsuperscript{BB} based on the analysis of the lay user explanation. }
    \label{fig:LayRelations}
\end{figure}
With these results, we could synthesize the expert's concepts and the user's mental models in a structured way.
\subsection{Synthesis}
\label{sec:Synthesis}
\begin{figure}
    \centering
    \includegraphics[width = \textwidth]{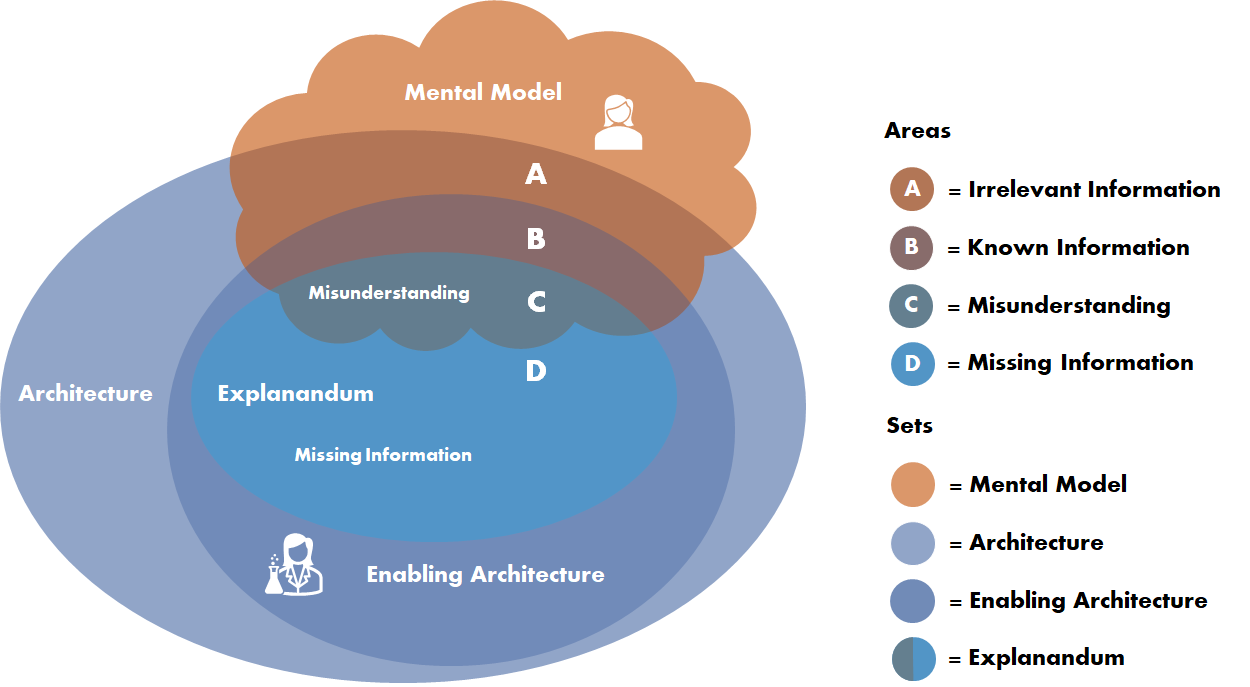}
    \caption{A representation of areas of knowledge regarding the architecture and the lay user.}
    \label{fig:Van-Daigram-Knowledge}
\end{figure}

\begin{figure}
    \centering
    \includegraphics[width= \textwidth]{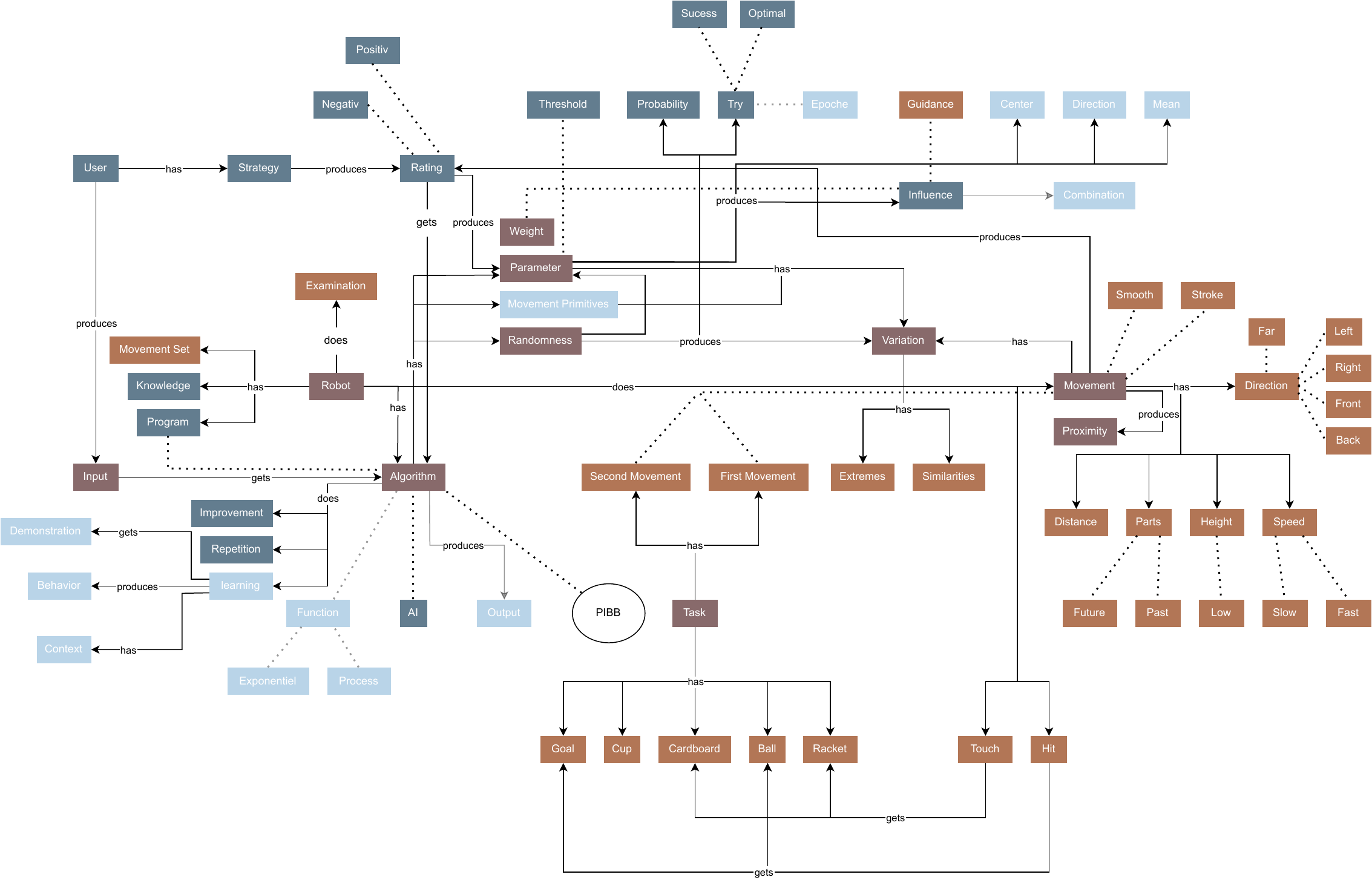}
    \caption{The classified version of the concept map of pupils' mental model.}
    \label{fig:LayColored}
\end{figure}

First, we have considered different types of knowledge that are given by the corresponding architecture and the lay users. 
To illustrate this, we have depicted it in a Van-diagram (Figure \ref{fig:Van-Daigram-Knowledge}), which we will explain in more detail below.
On the one hand, there is all the knowledge about the algorithm. 
In the context of the general approach, this would be architecture which is shown in light blue. 
Part of the architecture is the enabling architecture, shown in dark blue, which we extracted from the expert interviews. 
Next to this is the users' overall mental model, which is shown in orange.
This contains irrelevant information that is not needed in the context of the interaction marked in dark orange (see area A).
In addition, there is information that the user has already correctly stored in his mental model regarding the enabling architecture.
This is the already known information located in area B and illustrated with a blue-orange.
The part we focus on is the explanandum area D highlighted in turquoise.
This represents the part of the enabling architecture that still keeps the user from using the system efficiently.
The explanandum intersects the users mental model, which are concepts and interactions that the user already knows, but with the wrong assumption or in the wrong context. 
However, they are not irrelevant to the efficient usage of the architecture.
These are misunderstandings and represent area C marked blue-orange. 
Misunderstandings within the relevant architecture are thus part of the explanandum since they need to be clarified in order to allow efficient usage of the system.
Last but not least, there is area D representing the concepts and interactions the user is missing from the enabling architecture, which are part of the explanandum as well. 
The users need this information to effectively interact with the system. \\
Thus, the enabling architecture is part of the overall architecture and contains the information that a user might need to use the system effectively.
Part of the enabling architecture is the explanandum which is the information the user actually missing. 
It excludes the parts of the enabling architecture that are already known. 
The already-known information is part of the user's mental model and the enabling architecture. 
The explanandum itself is divided into misunderstandings and missing information.
The misunderstandings are part of the enabling architecture and the mental model of the user since they are information about the enabling architecture the user knows partially.
The missing information is not part of the mental model of the user but part of the enabling architecture.
The information that is not part of the enabling architecture, but part of the architecture, is considered irrelevant for the effective usage of the system.\\
Based on this subdivision, we have synthesized the diagrams of the experts and lay users.
This was done by classifying the concepts and interactions regarding classes A to D.
\begin{figure}
    \centering
    \includegraphics[width = 1 \textwidth]{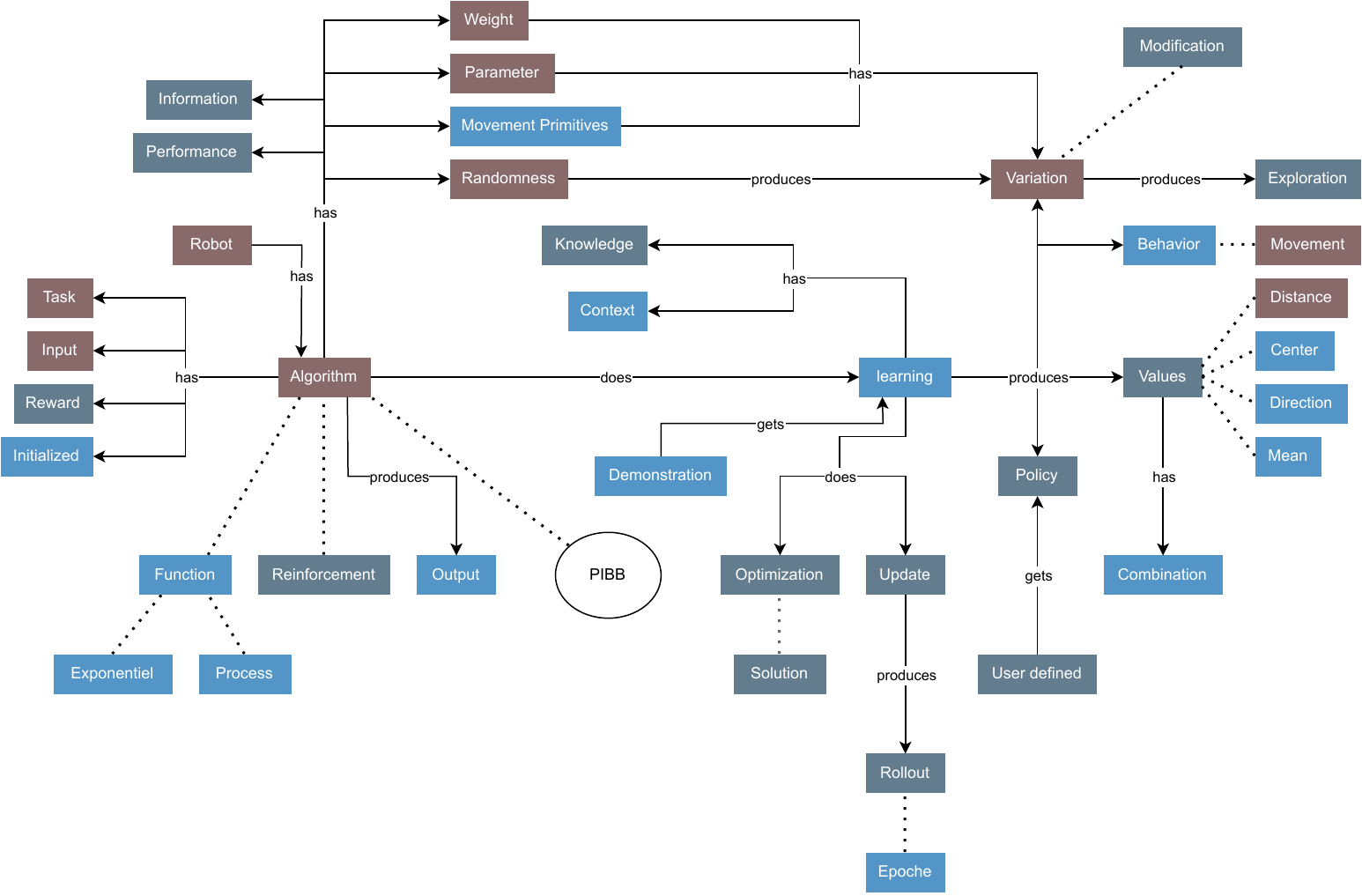}
    \caption{The classified version of the concept map of the enabling architecture.}
    \label{fig:ExpertColored}
\end{figure}
First, all concepts and interactions that appear in both concept maps depicted in Figure \ref{fig:ExpertRelations} and Figure \ref{fig:LayRelations} were marked in the same context and interpretation.
The concepts and interactions that were classified by this are part of area B the already-known information.\\ 
Afterward, the concepts covered in the wrong context or assumptions were classified as misunderstandings located in area C. 
For those concepts and interactions in the lay user diagram, a counterpart exists in the expert diagram. 
Otherwise, they would have been classified as irrelevant since they are then not part of the enabling architecture (Figure \ref{fig:ExpertColored}) or as missing since they would not exist in the lay user concept map (Figure \ref{fig:LayColored}).
This property of misunderstandings is natural since the lay user reasoned about the concept or interactions of the enabling, but as mentioned with wrong context or assumptions.\\
Last but not least, there are missing information and irrelevant information, 
both of which are exclusive to one type of concept map.
Only the lay user has irrelevant information, and the experts can only have the missing information.
Thus, all unclassified nodes in the two diagrams are then classified into these two classes accordingly.
For the following discussion of these results, we also marked the missing information in the lay user concept map where they would belong in transparent turquoise (c.f Figure \ref{fig:ExpertColored}). 
\section{Discussion}
\label{sec:Dicussion}
The structure of the discussion of the synthesis results of our example architecture is based on the areas of knowledge presented in Figure \ref{fig:Van-Daigram-Knowledge}. \\
% Auch bezug auf Task neben den sichtbaren dingen und unten Beispiel wie es bei uns war
We will first consider the area of irrelevant information (Area A).
The lay users tend to refer a lot to the actual task, which is irrelevant to the lay users understanding of the underlying algorithm. 
The algorithm can be applied to different tasks without changes to its functioning.
Thus, the experts explained the algorithm independent from the task, and the enabling architecture is task-independent. 
However, the algorithm is utilized in the context of a task.
Therefore, the lay users are introduced to the architecture in a task context.
This behavior is also evident in our study, where the lay users focus more on the physical environment in the form of 'Cup', 'Ball' or 'Goal' or individual characteristics of the robot arms stroke, such as the 'Direction', 'Speed' or 'Height'. 
% Diese Verhalten kann in unserer Beispiel Anwendung ebenfalls beobachtet werden. 
% Übergang zum nachfolgenden Satz evt etwas fließenden dann
% (c.f Figure \ref{fig:classified})
Also, the lay users often are focused on comparing the first and second stroke of a run and specify the differences between the individual movements in greater detail. 
This can be seen in the lay user concept map (c.f Figure \ref{fig:LayRelations}).\\
% This information regarding the study task is irrelevant in terms of comparison with the experts' specifications (c.f Figure \ref{fig:classified}), since the experts described  the underlying PI\textsuperscript{BB} algorithm without any reference to the study task performed by the lay users.
The following deals with the area of known information (Area B), including aspects mentioned by the lay users but also stated in the expert study.
For instance, the lay users have described how the robot has an algorithm with some kind of randomness and thus provides movement variation. 
Furthermore, the lay users also mentioned how the algorithm includes several parameters, which vary due to the specified randomness.
Additionally, the lay users report, like the experts, how the algorithm has various weights.
Both the experts and the lay users describe an input for the algorithm.
But, the experts explain how the algorithm has an input and how it is used, while the lay users describe that the algorithm gets input but do not delve into what happens with it afterward.\\
Our focus was on the explanandum, which consists of the misunderstandings (Area C) and the missing information (Area D).
We first cover the misunderstandings where concepts were only roughly understood correctly or were assigned to the wrong context.\\
%Allgemeine Erkenntinis bei den Missunderstandings
% In between, there were some misunderstandings among the lay users, where concepts were only roughly understood correctly or were assigned to the wrong context, which refers to the area of misunderstanding (Area C). 
%Often the rough understanding and wrong categorization resulted from 
This was also noticeable in our study.
Misinformation seems to arise when lay users are aware of the existence of a concept or interaction but lack understanding regarding its processing, utility, functionality, or context.
In contrast to the known information (Area B), those are concepts and interactions that are not derivable by the perceived situation or are less commonly known and thus not part of the prior knowledge of the lay users. 
For example, the existence of the rating was present since it was part of the task and user interface. However, its processing and utility as a reward in a reinforcement learning context were not derivable by the observation and not part of the lay users' prior knowledge.
Thus its existence was mentioned but was misunderstood since the core features of this concept are missing. 
Nevertheless, the user would probably benefit from a deeper understanding of the reward concept since it would allow for more precise manipulation of the learning process. 
% The term "knowledge", for instance, was attributed by the lay users to the robot, while the experts associated it with the learning process of the algorithm.
Furthermore, the term 'knowledge' was attributed to the robot by the lay users because they were unaware that the robot, in this case, did not possess actual knowledge, but rather the algorithm as described by the experts. This also demonstrates that the existence of knowledge was known, but the context and utility were misunderstood.\\
% Was unterscheidet die gewussten von den fast gewussten?
% Verarbeitung Nutzen Kontext<- zu ordnung zu was das gehört 
% Exsitenz ist da 
% Was fehlt den fast gewussten um gewusst zu sein?
% s.o.
% Warum sind sie nicht ungewusst was haben sie schon?
% weil Exsitenz bekannt aber nicht aus dem sichtbaren erschließbar und nicht gut abgedeckt von Prior Knowledge. Das das auch im Zusammenhang mit Abstraktionsgrad da je nachdem prior knowoldge. 
The second part of the explanandum is the missing information (Area D) covered next.
Mostly, the lay users lacked information about the operating principles of the robot.
Some concepts were not mentioned by the users at all, which could imply that the users are either unaware of their existence in this context or their existence in general. 
That these concepts and interactions are missing or not associated with this context could be related to the fact that they are less commonly known and thus not part of the prior knowledge of the users.
They are also not directly perceivable like 'Robot' or derivable like 'Randomness'. 
One example of a concept that the user probably did not know of its existence is 'Movement Primitive'.
'Mean', on the other hand, the participants most likely know but did not consider it as a core concept of the PI\textsuperscript{BB} algorithm. 
% which is mapped to the area of missing information (Area D).
Another important example of this mismatch of context is 'Learning'.
The lay users did not associate the 'Learning' process in the context of artificial intelligence.
They only referred to learning as a human-like learning process.
In addition, unlike the experts, the lay users did not talk about any 'Function' or 'Process' in relation to the algorithm.

\begin{figure}
    \centering
    \includegraphics[width= 0.8\textwidth]{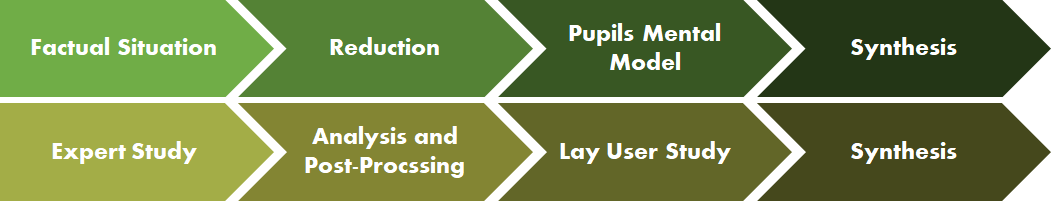}
    \caption{The steps of identifying the enabling architecture and the explanandum.}
    \label{fig:FlussDiagram}
\end{figure}

Overall, it is recognizable how the lay users have mostly referenced physically perceptible artifacts, like the movement of the robot arm. However, lay users have referred less to the underlying structures, like how the robot arm controls its movement.
The algorithm was more like a black box for the users.
They referred to the inputs and outputs as well as components like weights and randomness but did not reason about their interaction and causalities. \\
Overall, this classification, for example, could now be a base for the according explanation of the algorithm. 
The known information can be considered given and the misunderstandings would need to be clarified, and the missing information would need to be supplemented. Those explanations could then be used, for example, in a context like explainable artificial intelligence in robotics.\\
%%schlussfolgerung:
%%lay user haben eine leichte idee was passiert.
%%lay user vermsenschlcihen arm, und reden daher viel ueber die bewegung an sich und wie man sich eine normale bewegung vorstellt?
%%Es existieren einige Missverstaendnisse, vielleicht gegeben durch halbwissen, wo dann falsches denken ueber details entsteht. Beispiel welchen Einfluss hat Bewertung auf schlaege. 
In conclusion, we can summarize the identification of the enabling architecture and the corresponding explanandum (Figure \ref{fig:FlussDiagram}).
\begin{itemize}
    \item Factual Situation - Expert Study (Section \ref{sec:FactualSituation})\\
    First, the architecture needs to be identified. For this, the experts of the target architecture must be queried for explanations.
    \item Reduction - Analysis and Post-Procsssing (Section \ref{sec:architectureReduction})\\
    Second, the architecture retrieved needs to be reduced by performing the structured analysis.
    By removing all function words and merging different grammatical cases. 
    Afterward, the analysis results are post-processed, content-related duplications, and low-frequency concepts and interactions are removed. 
    The reduction can be visualized using a concept map. 
    \item Pupil's mental models - Lay User Study (Section \ref{sec:LayMentalModel})\\
    Third, the user's mental model of the architecture needs to be assessed.
    % For this, there exist different methods.
    For this, the lay users can be interviewed in the form of semi-structured video recall.
    Care should be taken to ensure that users are interviewed in such a way that the mental models queried are those that represent the concepts from the analysis of the expert explanations.
    Furthermore, lay users should not be biased toward certain concepts and interactions during the interview. 
    \item Synthesis (Section \ref{sec:Synthesis})\\
    Last, the experts' reduced architecture and the users' mental models must be synthesized.
    For this, the involved sets of knowledge must be considered. 
    These sets are the architecture, enabling architecture, and the user's mental model. 
    The overlap of those sets shapes the areas that are the basis of the synthesis. 
    On top is the explanandum, which is part of the enabling architecture.
    The overlap of the explanandum with the sets of knowledge and the overlap of the sets themselves result in the four areas that allow for the classification of the concepts and interactions within the experts enabling architecture and the user's mental model. 
    The explanandum can be identified based on the classification of the missing information and the misunderstandings and exclusion of the irrelevant and already known information.
\end{itemize}

\subsection{Future Work}
Questions that remain open include the influence of modalities on users' mental models and, thus, on the enabling architecture. 
We excluded them from this paper and only provided preference as a modality.
It stands to reason that modalities change the mental model and the enabling architecture. For example, scalar evaluations require users to create a global scalar that is valid across all interactions instead of preferences only valid within one interaction. \\ 
Another open question is the definition of the level of abstraction, which is a future part of the synthesis step of our approach.
The abstraction would define which concepts and interactions of the reduced architecture needs to be adapted to the level of abstraction the user needs. 
With abstraction, we refer to the level of complexity of the experts' concepts, which might need to be reduced to be accessible to lay users.
A first hint to the structured determination of the abstraction level can be derived from the different areas of knowledge since already-known information can be used as a base for the explanation.
The misunderstandings are concepts and interactions of which the users know their existence but do not know further details.
Therefore, those seem to be their minimum level of abstraction, and the missing information is below their level of abstraction. \\
Furthermore, as mentioned in the beginning, the form of the presentation is also an open question. 
This includes a possible explanation structure, which is not included in the approach.
However, we also work on analyzing the explanation structure of our example architecture.\\
Last but not least, the automation of our approach is future work.
The systematic approach allows using, for example, knowledge graph generations or large language models
%Zitat
for the reduction of the factual situation.



%%       \citet{<label>} ==> Jones et al. [21]
%%       \citep{<label>} ==> [21]
%%

% \section*{Acknowledgements}
% We acknowledge the financial support of the German Research Foundation (DFG) and Transregional Collaborative Research Centre TRR 318 "Constructing Explainability".
\section*{Funding}
\textit{This work was supported by German Research Foundation (DFG) and Transregional Collaborative Research Centre TRR 318 "Constructing Explainability".}\\

\textit{Mara Brandt gratefully acknowledges the financial support from SAIL. SAIL is funded by the Ministry of Culture and Science of the State of North Rhine-Westphalia under the grant no NW21-059A.}\\

\textit{Phillip Richter gratefully acknowledges the financial support from Honda Research Institute Europe for the project "Reducing Mental Model Mismatch for Cooperative Robot Teaching".}
\bibliographystyle{elsarticle-num-names} 
\bibliography{bibliography}





% \end{thebibliography}
\end{document}